%% file: main.tex
\documentclass[letterpaper, 10 pt, conference]{ieeeconf}  %

\IEEEoverridecommandlockouts                              %
\usepackage[T1]{fontenc}                                                          

\usepackage{tikz} %

\usepackage{amsfonts}	%
\usepackage{amsmath}	%
\usepackage{amssymb}    %
\usepackage{siunitx}
\usepackage{pifont}   %
\usepackage{xspace}  

\usepackage{svg}

\usepackage{booktabs}
\usepackage{makecell}  %
\usepackage[flushleft]{threeparttable}  %
\usepackage{multirow}
\usepackage{rotating}

\usepackage{xspace}    %
\usepackage{xcolor}    %
\let\labelindent\relax
\usepackage[inline]{enumitem} %
\usepackage{textcomp}  %
\usepackage{gensymb}   %
\usepackage{graphicx} %
\usepackage{microtype}
\usepackage{cite}
\usepackage{flushend}
\usepackage{cuted}

\usepackage{pgfplots}
\pgfplotsset{width=7cm, compat=1.10}
\usepgfplotslibrary{fillbetween}
\usepackage{tabularx}

\newcolumntype{P}[1]{>{\centering\arraybackslash}p{#1}}

\makeatletter
\let\NAT@parse\undefined
\makeatother

\usepackage{url}
\usepackage[pdfencoding=auto, breaklinks,colorlinks,hypertexnames=false]{hyperref} %
\usepackage[hang,flushmargin]{footmisc}

\usepackage[capitalize]{cleveref}
\crefname{section}{Sec.}{Secs.}
\Crefname{section}{Section}{Sections}
\Crefname{table}{Table}{Tables}
\crefname{table}{Tab.}{Tabs.}

\usepackage[nogroupskip,acronyms,nopostdot,style=super,nonumberlist,toc]{glossaries}

\usepackage{colortbl}

\newcommand{\ours}{KeySG\xspace}
\newcommand{\oursextended}{KeyFrame-Based 3DSGs\xspace}
\newcommand{\website}{\url{https://keysg-lab.github.io/}}
\newcommand{\m}[1]{\mathcal{#1}}

\input{preamble/math_preamble}

\input{preamble/text_preamble}

\title{\LARGE \bf
KeySG: Hierarchical Keyframe-Based 3D Scene Graphs
}

\author{%
  Abdelrhman Werby \quad
  Dennis Rotondi \quad
  Fabio Scaparro \quad
  Kai O. Arras \\
  \thanks{All the authors are with the Socially Intelligent Robotics Lab, Institute for Artificial Intelligence
  University of Stuttgart, Germany.
  Email: \texttt{\{first.last\}@ki.uni-stuttgart.de}. A. Werby and D. Rotondi are also part of the International Max Planck Research School for Intelligent Systems (IMPRS-IS).}%
}

\begin{document}

\maketitle
\thispagestyle{empty}
\pagestyle{empty}

\begin{abstract}
    \input{sections/0_abstract}
\end{abstract}
\glsresetall

\input{sections/1_introduction}
\input{sections/2_related_work}

\input{sections/3_approach}
\input{sections/4_experiments}

\input{sections/5_conclusion}

\begin{footnotesize}
    \bibliographystyle{IEEEtran}
    \bibliography{main.bib}
\end{footnotesize}

\end{document}

%% file: preamble/math_preamble.tex
\usepackage{amsmath}
\usepackage{amssymb}

\DeclareMathAlphabet{\mathcal}{OMS}{cmsy}{m}{n}

\def\R{\mathbb{R}}

%% file: preamble/text_preamble.tex
\usepackage{xspace}

\renewcommand{\[}{\begin{equation}}
\renewcommand{\]}{\end{equation}}

\renewcommand{\baselinestretch}{0.99}

\definecolor{red}{RGB}{255, 0, 0}   %
\definecolor{orange}{RGB}{255, 77, 0}   %
\definecolor{green}{RGB}{0, 128, 0}   %
\definecolor{purple}{RGB}{160, 32, 240}   %
\definecolor{lightblue}{RGB}{52, 155, 235}   %
\definecolor{darkmagenta}{RGB}{204, 51, 139} %

\usepackage[capitalize]{cleveref}

\crefname{figure}{Fig.}{Figs.}
\Crefname{figure}{Figure}{Figures}
\crefname{section}{Sec.}{Secs.}
\Crefname{section}{Section}{Sections}
\Crefname{table}{Table}{Tables}
\crefname{table}{Tab.}{Tabs.}
\crefname{algorithm}{Algo.}{Algos.}
\Crefname{algorithm}{Algorithm}{Algorithms}
\crefname{appendix}{Sec.}{Secs.}
\Crefname{appendix}{Section}{Sections}

%% file: sections/0_abstract.tex
In recent years, 3D scene graphs have emerged as a powerful world representation, offering both geometric accuracy and semantic richness.
Combining 3D scene graphs with large language models enables robots to reason, plan, and navigate in complex human-centered environments.
However, current approaches for constructing 3D scene graphs are semantically limited to a predefined set of relationships, and their serialization in large environments can easily exceed an LLM's context window.
We introduce \ours, a framework that represents 3D scenes as a hierarchical graph consisting of floors, rooms, objects, and functional elements, where nodes are augmented with multi-modal information extracted from keyframes selected to optimize geometric and visual coverage.
The keyframes allow us to efficiently leverage VLMs to extract scene information, alleviating the need to explicitly model relationship edges between objects, enabling more general, task-agnostic reasoning and planning. Our approach can process complex and ambiguous queries while mitigating the scalability issues associated with large scene graphs by utilizing a hierarchical multi-modal retrieval-augmented generation (RAG) pipeline to extract relevant context from the graph. 
Evaluated across three distinct benchmarks, 3D object semantic segmentation, functional element segmentation, and complex query retrieval \ours outperforms prior approaches on most metrics, demonstrating its superior semantic richness and efficiency. See our project page at \website

%% file: sections/1_introduction.tex
\section{Introduction}
\label{sec:introduction}
A long-standing goal in robotics is to create autonomous agents that can operate effectively in human-centered environments such as homes or offices. 
These environments are characterized by high object density, semantic richness, and a variety of potential tasks. 
A key challenge for this goal is the development of a 3D world representation that is simultaneously detailed for precise manipulation and abstract enough for high-level reasoning and long-horizon planning. 

\begin{figure}[t!]
    \centering
    \includegraphics[width=1.0\linewidth]{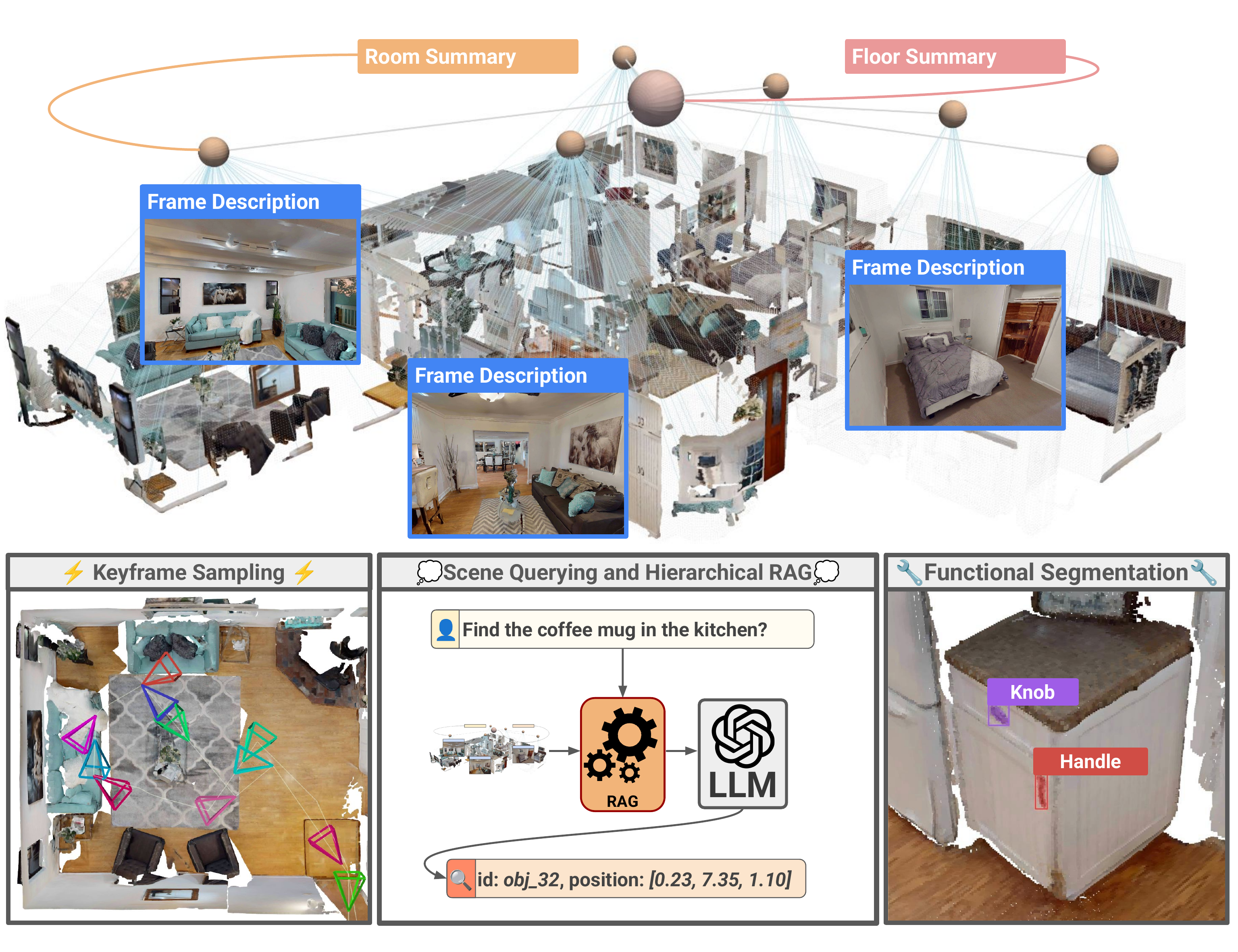}
    \caption{As illustrated (top), \ours is a hierarchical, keyframe-based 3D scene graph comprising floors, rooms, objects, and functional elements (bottom right). Each node is augmented with contextual information efficiently extracted from scene keyframes via keyframe sampling (bottom left). Leveraging a multimodal RAG pipeline, \ours enables users to ask complex natural language queries and receive answers grounded in the 3D scene (bottom middle).}
    \label{fig:cover}
    \vspace{-.7cm}
\end{figure}

3D scene graphs (3DSGs)~\cite{conceptgraphs, rosinol20203DDS, hughes2022hydra, wald2020learning3dsemanticscene, rana2023sayplan} have gained significant attention as a powerful representation to address the limitations of purely geometric maps. By modeling the world as a graph where nodes represent entities and edges represent their relationships, 3DSGs impose a structure on raw perception, explicitly linking geometry to semantics. 
However, current 3D scene graph approaches have two main limitations: first, they are restricted to a predefined set of geometric or semantic relationships, reducing the diversity of tasks and queries they can support. 
For instance, a 3DSG \cite{Werby-RSS-24} with edges representing spatial relationships between objects and places would excel in locating objects in large buildings. In contrast, a 3DSG \cite{rotondi2025fungraph} encoding functional relationships would be well suited for tasks that require understanding how functional elements control their objects (e.g, ``turn off the oven," where knowing the relationship between the knob and the oven is necessary). A 3D scene graph designed with a predefined set of relationships for a specific task is inherently suboptimal for others.

Second, scalability is a major bottleneck.  3D scene graphs are often paired with large language models (LLMs), serving as a persistent world model that the LLM uses for planning and high-level reasoning. However, providing a complete, detailed scene graph of a large-scale environment, such as a multi-story office building, directly to an LLM can exceed the context window limits of even the most advanced models.
Even if the graph fits within the context window, LLMs suffer from attentional biases and a ``lost in the middle" problem ~\cite{hsieh2024found}, where performance degrades as the model is distracted by the vast amount of task-irrelevant information present in the prompt. This makes it challenging for the LLM to identify the crucial entities and environmental states necessary for robust reasoning and planning.

To this end, we present the Hierarchical KeyFrame-Based 3D Scene Graphs (\ours), a novel framework that resolves the semantic and scalability dilemma by augmenting the 3D scene graph with multi-modal contextual information, implicitly capturing the geometry, semantics, affordances, and states of objects. Our key idea is to sample keyframes that ensure comprehensive visual coverage of each room in the environment. A Vision-Language Model (VLM) then generates a detailed description for each keyframe. To address scalability, these descriptions are recursively summarized into concise textual overviews for rooms and, subsequently, entire floors. This hierarchy is queried using a multi-modal retrieval-augmented generation (RAG) pipeline, which ensures that only the most relevant information is retrieved and provided to the LLM planner, enabling efficient and accurate reasoning.

In summary, we make the following contributions:
\begin{itemize}
    \item We introduce \oursextended (\ours), the first 3D scene graph framework designed to hierarchically represent environments spanning from full buildings, to floors, rooms, objects, down to functional elements.
    \item We propose a new pipeline to augment 3DSGs with multi-modal context from keyframes, featuring hierarchical scene summarization, and a RAG-based retrieval mechanism that efficiently provides task-relevant context to LLMs. 
    \item We conduct a comprehensive evaluation of \ours\ across diverse benchmarks, including open-vocabulary 3D segmentation (Replica \cite{straub2019replica}), functional element segmentation (FunGraph3D \cite{zhang2025open}), and 3D object grounding from natural language queries (Habitat Matterport \cite{habitat23semantics}, Nr3D \cite{achlioptas2020referit_3d}).
\end{itemize}

%% file: sections/2_related_work.tex
\section{Related Work}
\label{sec:related_work}

\subsection{3D Scene Graphs}

The idea of 3D Scene Graphs (3DSGs)~\cite{armeni20193d, kim2019sparse3d} is to represent a scene as a graph \(\m{G} = (\m{V}, \m{E})\), where the nodes \(\m{V}\) correspond to objects and spatial entities (e.g., cameras, rooms, floors, buildings), while the edges \(\m{E}\) encode relationships such as \texttt{hierarchical} (e.g., A ``is part of'' B), \texttt{spatial} (e.g., A ``is next to'' B), and \texttt{comparative} (e.g., A ``is larger than'' B).
Unlike geometric maps fused with language features~\cite{shah2023lm, conceptfusion, yamazaki2024open}, 3DSGs provide higher-level abstraction, scale naturally to large environments~\cite{hughes2024foundations}, and, thanks to detailed node captions and semantic relationships, have proved useful in robotics tasks such as planning~\cite{agia2022taskography, rana2023sayplan, bartoli2025, rotondi2025social3dsg}, manipulation~\cite{momallm24, yan2025dynamicopenvocabulary}, and navigation~\cite{conceptgraphs, Werby-RSS-24}.

3DSGs are typically constructed either from dense sequences of RGB-D images~\cite{rosinol2020dynamic, rosinol2021kimera, hughes2022hydra} or from class-agnostic segmented point clouds~\cite{wald2020learning3dsemanticscene, wu2021scenegraphfusionincremental3dscene, koch2024open3dsgopenvocabulary3dscene}.

In general, all available input is used to construct the graph, and in some approaches, images that meet specific criteria --for example, those focusing on a particular pair of objects~\cite{conceptgraphs, koch2024open3dsgopenvocabulary3dscene, linok2025beyondbarequeries}-- are used to establish spatial edges, while others are selected to extract functional interactive relationships~\cite{rotondi2025fungraph, zhang2025open} (e.g., a button ``turns on" a monitor).
In contrast to our approach, all these solutions explicitly model edges and disregard the RGB input once the graph is constructed, thereby constraining the 3DSG to the specific application for which it was built. %

To address this problem, the works of \cite{Maggio2024Clio, yun2025ashita} attempt to reason about object primitives at the resolution required for a specific task by mitigating the information bottleneck. However, a key limitation of these methods is that even a slight change in the task requires reconstructing the entire graph from scratch. In contrast, \cite{koch2025relationfield} generates a 3DSG from a NeRF, capturing the necessary relationships between pairs of objects, but these relationships are restricted to those modeled by the NeRF at training time. 
Our method overcomes both issues by simply rendering new edges from our keyframes to answer specific queries.

\subsection{Object Localization with 3D Scene Graphs}
3D object localization \cite{chen2020scanrefer}, also known as object grounding, is the task of predicting a target 3D object given a point cloud and a natural language expression as input.
Contrary to other approaches~\cite{llmgrounding, OpenEQA2023, huang2024chat}, 3DSGs do not need to be trained on ground-truth point clouds, nor do they require fine-tuning of the LLM itself for object grounding.
In practice, object localization with 3DSGs is often realized by serializing the graph into JSON \cite{rana2023sayplan, conceptgraphs ,rotondi2025fungraph} and prompting an LLM to find the target object using the graph structure and node features. Different works have tried to optimize the JSON by adding hierarchical structure \cite{momallm24}, using the concept of schema \cite{loo2025open} or by applying a deductive scene reasoning algorithm \cite{linok2025beyondbarequeries} that prunes the search space using spatial relationships.
All these methods are limited to the information contained in the 3DSG, which is often insufficient to disambiguate complex queries that include specific edges.

%% file: sections/3_approach.tex
\section{Technical Approach}
\begin{figure*}[t]
    \centering
    \includegraphics[width=1.0\textwidth,keepaspectratio]{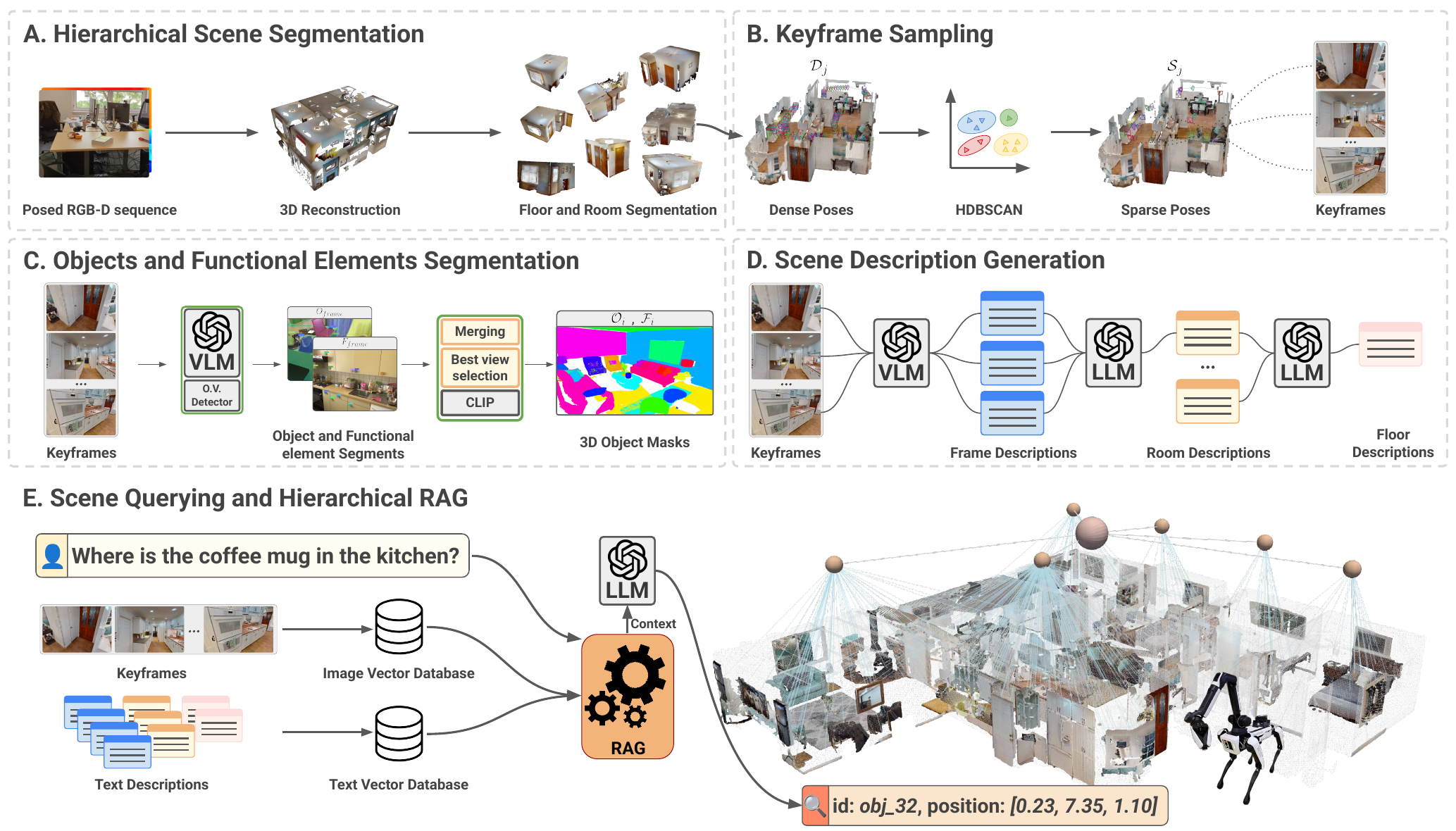}
    \vspace*{-.5cm}
    \caption{Overview of \ours: (A) We first reconstruct the full point cloud of the 3D scene and segment it into floors and rooms. (B) For each room, we select a minimum set of keyframes that provide geometric coverage of the entire space while maximizing visual information. (C) We leverage VLMs to extract object and functional element tags from the keyframes, which guide an open-vocabulary segmentation pipeline to obtain 3D segments of objects and functional elements. (D) We generate geometrically-grounded frame descriptions using a VLM, and employ LLMs to recursively summarize them into dense room and floor-level overviews. (E) To enable efficient querying, we introduce a hierarchical retrieval mechanism grounded in RAG that performs a top-down search, ensuring LLMs receive rich, task-relevant context without exceeding their context window.
    }
    \label{fig:overview}
    \vspace*{-.4cm}
\end{figure*}

This work aims to build a hierarchical 3D scene graph for a large-scale environment, consisting of floors, rooms, objects, and functional elements, where each node is augmented with multi-modal knowledge extracted from keyframes in the environment.
Given a posed RGB-D image sequence, we first reconstruct a dense 3D point cloud of the environment and segment it into floors and rooms (Sec.~\ref{subsec:scene_segmentation}). Within each room, we select keyframes that maximize both geometric coverage and visual informativeness (Sec.~\ref{subsec:keyframe_sampling}). A VLM then extracts textual descriptions, object labels, and functional element labels, which guide an open-vocabulary segmentation pipeline to produce 3D object and functional element segments (Sec.~\ref{subsec:obj_fun_seg}). An LLM subsequently condenses keyframe descriptions into room summaries and aggregates them into floor-level summaries, thereby building contextual information across multiple levels of abstraction within the 3DSG (Sec.~\ref{subsec:scene_desc}). Finally, we introduce a multi-modal hierarchical Retrieval-augmented generation (RAG) pipeline inspired by~\cite{edge2024local}, which exploits the graph's topology to support top-down querying, ensuring that LLMs receive task-relevant information without exceeding their context window (Sec.~\ref{subsec:rag}). Fig.~\ref{fig:overview} provides an overview of our approach.

\subsection{Hierarchical Scene Segmentation}
\label{subsec:scene_segmentation}

Given a posed RGB-D sequence $\mathcal{I}=\{P_t, I_t, D_t\}_{t=1}^T$ of the environment, where $P_t = [R_t | t_t] \in SE(3)$ is the camera pose, $I_t$ is the RGB image, and $D_t$ is the depth map, we first reconstruct a global 3D point cloud  $\mathcal{P}_{scene} \in \R^{N \times 3}$ of the entire scene. Second, we segment the full scene point cloud $\mathcal{P}_{scene}$ into a set of $N_F$ floor point clouds, $\{\mathcal{F}_i\}_{i=1}^{N_F}$. Then we segment each floor point cloud $\mathcal{F}_i$ into a set of $N_{R_i}$ room point clouds $\{\mathcal{R}_{ij}\}_{j=1}^{N_{R_i}}$. This hierarchical segmentation follows the approach from \cite{Werby-RSS-24}, segmenting floors by computing a height histogram of the point cloud, then selecting dominant peaks. For room segmentation, we compute a 2D histogram from the bird's-eye view of the floor and apply the Watershed algorithm.

\subsection{Keyframe Sampling}
\label{subsec:keyframe_sampling}

A key principle of \ours is to augment graph nodes with both raw sensory data and semantic context using VLMs.
However, storing and processing the entire data sequence for a large-scale environment is computationally impractical.
To address this, \ours employs a down-sampling procedure that balances computational efficiency with the preservation of critical spatial and visual information. While this problem is known from keyframe-based visual SLAM~\cite{YOUNES201767}, we select frames based on visual room coverage rather than geometric reconstruction accuracy.

We start by associating each frame from the original sequence with a specific segmented room $\mathcal{R}_{ij}$ (Room $j$ in Floor $i$).
A frame's camera pose is assigned to a room if its associated 3D camera center falls within the room's volumetric boundary, i.e., $t_t \in \mathrm{Vol}(\mathcal{R}_{ij})$.
This process yields a dense set of poses for each room $\mathcal{R}_{ij}$, denoted by $\mathcal{D}'_{ij} = \{P_t \in \mathcal{I} \mid t_t \in \mathrm{Vol}(\mathcal{R}_{ij})\}$.
To avoid corner cases where the frame's camera pose lies in one room but is viewing a different room or corridor, we further filter $\mathcal{D}'_{ij}$ by requiring that a significant fraction $\eta$ of the frame's back-projected 3D points also fall within the room's 2D polygon. The resulting refined set is denoted by $\mathcal{D}_{ij}$.

Next, we extract a subset $\mathcal{S}_{ij} \subseteq \mathcal{D}_{ij}$ such that $|\mathcal{S}_{ij}| \ll |\mathcal{D}_{ij}|$.
For each pose matrix $P_t \in \mathcal{D}_{ij}$, we extract pose features as 7D vectors $f_t = (t_t; w \cdot q_t)$, where $q_t \in \mathbb{R}^4$ is the quaternion derived from its rotation matrix $R_t$ and $w$ is a scalar rotation weight.
The features are then standardized:
\begin{equation}
\tilde{f}_t = \frac{f_t - \mu}{\sigma}
\end{equation}
where $\mu$ and $\sigma$ are the mean and standard deviation computed across all pose features.
We apply HDBSCAN~\cite{hdbscan} clustering to the set of standardized features, utilizing a minimum cluster size of 15 and a rotation weight $w=1.5$. This yields a set of clusters $\mathcal{C}_{ij}$.
For each cluster $c_k \in \mathcal{C}_{ij}$, we compute the medoid, i.e.,
the point $f_{k}^* \in c_k$ that minimizes the sum of distances to all other points within that cluster:
\begin{equation}
f_{k}^* = \arg\min_{f_j \in c_k} \sum_{f_t \in c_k} \|\tilde{f}_j - \tilde{f}_t\|_2.
\end{equation}
The final set of selected keyframes, $\mathcal{S}_{ij} = \{f_k^* \mid k = 1, \dots, |C_{ij}|\}$, is formed by collecting the medoid from each cluster.

\subsection{Objects and Functional Elements Segmentation}
\label{subsec:obj_fun_seg}

After obtaining the representative keyframes for each room, the pipeline proceeds with the 3D segmentation of objects and their corresponding functional elements.
For the keyframes $\mathcal{S}_{ij}$ extracted from each room $\mathcal{R}_{ij}$, we first utilize a VLM~\cite{gpt4} to generate a list of object tags and functional element tags for every keyframe.
These lists are then aggregated and deduplicated across all keyframes to form the comprehensive tag sets $\mathcal{O}_{ij}$ (objects) and $\mathcal{E}_{ij}$ (functional elements) for the entire room.
Next, we use open-vocabulary object detection and segmentation models~\cite{fu2025llmdet, kirillov2023segment}, guided by the object tags $\mathcal{O}_{ij}$, to perform 3D segmentation on the keyframe set of room frames $\mathcal{S}_{ij}$.
This yields a set of redundant point clouds in global coordinates for each object.
To consolidate these, we incrementally merge objects with significant geometric overlap, following the approach in~\cite{Werby-RSS-24}.
Each merged object incorporates 2D masks from multiple viewpoints.
For deriving a canonical semantic feature, we apply a best-view selection strategy: each 2D mask is scored based on its size and distance from the image boundaries, prioritizing large, centrally located views.
The highest-scoring view is then used to compute a CLIP~\cite{tschannen2025siglip} embedding, ensuring a high-quality feature representation.
Finally, we extract the 3D segments of associated functional elements using the best view from each object, leveraging open-vocabulary models~\cite{fu2025llmdet, kirillov2023segment} with the functional element tags $\mathcal{E}_{ij}$.

\subsection{Scene Description Generation}
\label{subsec:scene_desc}   
In \ours, we argue that scene details, such as layout, semantics, object relationships, state, and affordance, are implicitly stored within the keyframes and their corresponding text descriptions, alleviating the need to explicitly model these specific relations as edges in the 3D scene graph. In this step, we utilize the sparse room keyframes and a VLM to generate a detailed description of each frame. To force the VLM to generate a geometrically-grounded frame description, we determine if an object point cloud is visible from a keyframe's perspective, and we perform a multi-step visibility check. First, object points are transformed into the camera's coordinate system. These points are then projected onto the keyframe 2D image plane, and any points falling outside the image are culled. For the remaining points, we perform an occlusion check by comparing each point's depth to the corresponding value in the keyframe depth map. The object is considered visible if the fraction of its visible points exceeds a minimum threshold $\theta_{\text{vis}}$. We feed the VLM with a keyframe image and the list of objects that are visible in it, resulting in a description that is geometrically grounded to the 3D objects we found in the scene. We then aggregate all keyframe descriptions to generate a comprehensive summary for each room. Subsequently, we aggregate all room summaries to generate the overall floor summary.

To construct the final hierarchical 3D scene graph, we assign each extracted context entity to its appropriate level based on spatial and semantic containment. Floors form the top-level nodes, encapsulating aggregated floor summaries. Within each floor node, room nodes are nested, each containing the segmented room's point cloud, keyframe multi-modal data, and room summary. Objects are assigned as child nodes under their respective rooms, incorporating their 3D merged point clouds, CLIP embeddings, brief descriptions, and associated functional elements as sub-nodes. The \ours hierarchy reflects the environment's physical organization, and it is semantically rich, making it applicable for a wide range of tasks.

\subsection{Scene Querying and Hierarchical RAG}
\label{subsec:rag}

In general, 3D scene graphs are often paired with LLMs~\cite{conceptgraphs, Werby-RSS-24, rotondi2025fungraph} to answer user queries. However, serializing the entire scene graph into a single prompt can overwhelm an LLM's context window. Moreover, LLMs' performance degrades when processing long contexts, making it harder for the model to retrieve relevant information. These issues hinder scalability to large environments.

To address this, we design a hierarchical multi-modal Retrieval-Augmented Generation (RAG) pipeline aligned with the structure of \ours.
This pipeline enables users to query the environment using \ours (e.g., ``Where is the coffee mug in the kitchen?") and receive answers grounded in the recovered 3D geometry and semantic data. It also supports indirect references via attributes, status, or spatial relations to other objects in the scene.
We initialize our RAG system through three main indexing steps: first, we create text chunks from all information extracted from \ours keyframes and group them by their graph level, yielding four chunk types: floor, room, frame, and object. Each chunk contains text as described in Sec.~\ref{subsec:scene_desc}. Second, we compute vector embeddings for all chunks and index them by type, using OpenAI's embedding models~\cite{gpt4} and FAISS~\cite{douze2024faiss}. Third, we build a visual vector database over all keyframes in the scene, as well as a separate database of object-level CLIP~\cite{tschannen2025siglip} embeddings.
Grounding a target can be challenging for complex queries, particularly those involving indirect references, attribute descriptions, or implicit targets. To handle this, we first employ an LLM to decompose the raw query $q$ into a set of semantic entities: \texttt{<target floor>}, \texttt{<target room>}, \texttt{<target object>}, and \texttt{<anchor objects>}. These entities guide the retrieval of a context bundle $C^*$ comprising a floor $F_i$, a room $R_{ij}$, and relevant contents $\mathcal{O}_{\text{rel}}, \mathcal{S}_{\text{rel}}$ that maximizes the posterior probability
\begin{equation*}
\small
\begin{aligned}
P(C|q) & \approx \underbrace{P(F_i | q) P(R_{ij} | F_i, q)}_{\text{Location Prior}} \\
& \prod_{o \in \mathcal{O}_{\text{rel}}} \underbrace{P(o | R_{ij}, q)}_{\text{Object}} \prod_{s \in \mathcal{S}_{\text{rel}}} \underbrace{P(s | R_{ij}, q)}_{\text{Visual}}
\end{aligned}
\end{equation*}
We approximate these conditional probabilities using cosine similarity between the query and the entity embeddings. We perform a top-$k$ selection at each level of the hierarchy, filtering objects and keyframes by the identified room $R_{ij}$ to ensure geometric consistency. The retrieved context $C^*$ can be supplied to an LLM to answer object-grounding or general 3D scene question-answering tasks.

%% file: sections/4_experiments.tex
\section{Experimental Evaluation}
In this section, we evaluate \ours on four different benchmarks to demonstrate both the quality of the geometric and semantic data stored in the graph nodes and its ability to accurately handle complex queries without explicit relationship modeling.
First, we assess its open-vocabulary 3D semantic segmentation capabilities against recent methods on the Replica dataset (Sec.~\ref{subsec:obj_sem_Seg}). 
Second, we compare it to recent approaches in functional element segmentation on the FunGraph3D dataset (Sec.~\ref{subsec:fun_sem_Seg}). 
Third, we evaluate its ability to retrieve objects from hierarchical queries in large-scale indoor environments using the Habitat Matterport 3D Semantic Dataset (Sec.~\ref{subsec:obj_ret_hier}). 
Finally, we examine its capabilities to ground objects from complex natural language queries that require understanding of the scene and its objects' shape, location, color, and affordance, using the Nr3D dataset (Sec.~\ref{subsec:obj_ret_com}).

\subsection{Open-vocabulary 3D Semantic Segmentation}
\label{subsec:obj_sem_Seg}
To evaluate the objects' visual semantic embeddings generated by \ours, we utilized scenes \texttt{office0-office4} and \texttt{room0-room2} from the Replica dataset~\cite{straub2019replica} to facilitate comparison with other researchers.
We followed the evaluation protocol from~\cite{conceptgraphs}: first, we extracted all semantic class names and modified them to ``an image of \{class name\}". Then, we computed the CLIP text embeddings for each class. Finally, we calculated the cosine similarity between these text embeddings and the object embeddings in the scene, assigning each object the class with the maximum similarity. We report mAcc as the class-mean recall and f-mIOU as frequency-weighted mean intersection over union (IoU). 
Tab.~\ref{tab:semseg} shows that \ours surpasses all recent approaches with a notable margin, demonstrating the effectiveness of extracting the visual CLIP embedding from the best object view.

\begin{table}[!t]
    \footnotesize
    \centering
    \renewcommand{\arraystretch}{1.05}
    \caption{} \vspace{-8pt}
    \setlength{\tabcolsep}{4pt}
    \begin{tabular}{llrr}
        \toprule
        & \textbf{Method} & mAcc & F-mIoU \\ 
        \midrule
        & Mask2former \cite{cheng2021mask2former} + CLIP \cite{radford2021learning} & 10.42 & 13.11 \\
        & ConceptFusion~\cite{conceptfusion} & 24.16 & 31.31 \\
        & ConceptFusion~\cite{conceptfusion} + SAM~\cite{kirillov2023segment} & 31.53 & 38.70 \\
        & ConceptGraphs ~\cite{conceptgraphs} & \underline{40.63} & 35.95 \\
        & ConceptGraphs-Detector~\cite{conceptgraphs} & 38.72 & 35.82 \\
        & Clio ~\cite{Maggio2024Clio} & 37.95 & 36.26 \\
        & HOV-SG ~\cite{Werby-RSS-24} & 38.07 & \underline{40.16} \\
        & \textbf{\ours (ours)} & \textbf{45.81} & \textbf{46.16} \\
        \bottomrule
    \end{tabular}
    {\begin{flushleft}
        Results for open-vocabulary 3D semantic segmentation of Replica dataset ~\cite{straub2019replica}. We report the mAcc and F-mIoU metrics (\%).
    \end{flushleft}}
    \label{tab:semseg}
\end{table}

\begin{table*}[!ht]
    \footnotesize
    \centering
    \renewcommand{\arraystretch}{1.05}
    \caption{} \vspace{-8pt}
    \setlength{\tabcolsep}{4.7pt} 
    \begin{tabular}{lccccccccc} 
        \toprule
        & \multicolumn{3}{c}{R@3} & \multicolumn{3}{c}{R@5} & \multicolumn{3}{c}{R@10} \\
        \cmidrule(lr){2-4} \cmidrule(lr){5-7} \cmidrule(lr){8-10}
        \textbf{Method} & IoU$_{> 0.0}$ & IoU$_{\ge 0.10}$ & IoU$_{\ge 0.25}$ & 
        IoU$_{> 0.0}$ & IoU$_{\ge 0.10}$ & IoU$_{\ge 0.25}$ & 
        IoU$_{> 0.0}$ & IoU$_{\ge 0.10}$ & IoU$_{\ge 0.25}$ \\
        \midrule
        OpenFunGraph \cite{zhang2025open} & \underline{45.34} & 5.39 & 0.31 & \underline{47.74} & 6.89 & 1.50 & \textbf{60.40} & 9.30 & 1.50 \\
        FunGraph \cite{rotondi2025fungraph} & 33.56 & \underline{22.03} & \underline{13.04} & 35.79 & \underline{22.93} & \textbf{13.64} & 39.98 & \underline{24.28} & \underline{14.30} \\
        \textbf{\ours (ours)} & \textbf{46.44} & \textbf{24.23} & \textbf{13.33} & \textbf{53.06} & \textbf{25.19} & \textbf{13.64} & \underline{57.12} & \textbf{27.57} & \textbf{14.53} \\
        \bottomrule
    \end{tabular}
    {\begin{flushleft}
        Results for 3D functional elements segmentation on the FunGraph3D dataset. Recall (R) grouped by Top-k and IoU thresholds metrics (\%) are reported.
    \end{flushleft}}
    \label{tab:recall_results}
\end{table*}

\begin{table*}[ht!]
    \footnotesize
    \centering
    \renewcommand{\arraystretch}{1.05}
    \caption{} \vspace{-8pt}
    \resizebox{\textwidth}{!}{
    \begin{tabular}{ll ccc ccc ccc}
    \toprule
     & & \multicolumn{3}{c}{R@1} & \multicolumn{3}{c}{R@5} & \multicolumn{3}{c}{R@10} \\
    \cmidrule(lr){3-5} \cmidrule(lr){6-8} \cmidrule(lr){9-11}
    \textbf{Method} & \textbf{Query Type} & IoU$_{> 0.0}$ & IoU$_{\ge 0.10}$ & IoU$_{\ge 0.5}$ & IoU$_{\ge 0.0}$ & IoU$_{\ge 0.10}$ & IoU$_{\ge 0.5}$ & IoU$_{> 0.0}$ & IoU$_{\ge 0.10}$ & IoU$_{\ge 0.5}$ \\
    \midrule
    \multirow{2}{*}{HOV-SG~\cite{Werby-RSS-24}} & (r, o) & 23.30 & 0.60 & 0.00 & 44.30 & 2.00 & 0.00 & 55.90 & 4.50 & 0.00 \\
     & (f, r, o) & 22.80 & 0.60 & 0.00 & 44.90 & 0.20 & 0.00 & 56.60 & 4.30 & 0.00 \\
    \midrule
    \multirow{2}{*}{\ours w/o RAG} & (r, o) & 32.62 & 26.50 & 15.50 & \textbf{70.75} & 61.25 & 40.05 & \textbf{83.25} & \textbf{75.50} & 53.00 \\
     & (f, r, o) & \textbf{35.30} & \textbf{30.37} & 15.80 & \underline{69.60} & \underline{61.90} & 40.30 & \textbf{83.10} & 75.00 & 51.50 \\
    \midrule
    \multirow{2}{*}{\ours w/ RAG} & (r, o) & \underline{34.00} & \textbf{30.40} & \textbf{20.60} & 68.10 & \textbf{62.00} & \textbf{45.90} & \underline{80.80} & \underline{75.10} & \textbf{58.30} \\
     & (f, r, o) & 32.90 & \underline{28.50} & \underline{18.40} & 68.00 & 61.00 & \underline{43.40} & 79.80 & 73.10 & \underline{55.50} \\
    \bottomrule
    \end{tabular}
    }
    {\begin{flushleft}
        Results for hierarchical 3D object retrieval on a large-scale environment within the HM3DSem dataset~\cite{habitat23semantics}. The evaluation spans 345 object categories with 2,809 queries per type. We report Recall (R) grouped by Top-k and IoU thresholds metrics (\%).
    \end{flushleft}}
    \label{tab:obj_ret_hier}
\end{table*}

\begin{table*}[ht!]
    \footnotesize
    \centering
    \renewcommand{\arraystretch}{1.05}
    \caption{} \vspace{-8pt}
    \resizebox{\textwidth}{!}{
    \begin{tabular}{l ccc ccc ccc}
    \toprule
      \multicolumn{1}{c}{} &
      \multicolumn{1}{c}{Overall} &
      \multicolumn{1}{c}{w Spatial Lang.} &
      \multicolumn{1}{c}{w/o Spatial Lang.} &
      \multicolumn{1}{c}{w Color Lang.} &
      \multicolumn{1}{c}{w/o Color Lang.} &
      \multicolumn{1}{c}{w Shape Lang.} &
      \multicolumn{1}{c}{w/o Shape Lang.} &
      \multicolumn{1}{c}{w Target Mention} &
      \multicolumn{1}{c}{w/o Target Mention} \\
    \textbf{Method} & IoU$_{\ge 0.10}$ & IoU$_{\ge 0.10}$ & IoU$_{\ge 0.10}$ & IoU$_{\ge 0.10}$ & IoU$_{\ge 0.10}$ & IoU$_{\ge 0.10}$ & IoU$_{\ge 0.10}$ & IoU$_{\ge 0.10}$ & IoU$_{\ge 0.10}$ \\
    \midrule
    OpenFusion~\cite{yamazaki2024open}  & 10.7 & 8.9 & 22.3 & 11.8 & 10.5 & 9.8 & 10.9 & 11.3 & 4.9 \\
    ConceptGraphs~\cite{conceptgraphs}  & 16.0 & 15.0 & 22.3 & 17.6 & 15.7 & 10.8 & 16.9 & 16.9 & 6.6 \\
    BBQ~\cite{linok2025beyondbarequeries}  & 28.3 & 28.1 & 29.8 & 25.2 & 29.0 & 34.3 & 27.3 & 29.6 & 14.8 \\
    KeySG w/ BBQ~\cite{linok2025beyondbarequeries}  & \underline{44.1} & \underline{45.8} & \underline{43.0} & \underline{46.5} & \underline{45.2} & \underline{28.9} & \underline{48.2} & \underline{47.5} & \underline{23.7} \\
    \textbf{KeySG w/ RAG (ours)} & \textbf{49.8} & \textbf{49.2} & \textbf{57.0} & \textbf{48.3} & \textbf{50.7} & \textbf{50.0} & \textbf{50.3} & \textbf{52.6} & \textbf{25.4} \\
  \bottomrule
  \end{tabular}
    }
    {\begin{flushleft}
        Results for 3D object grounding on the Nr3D dataset~\cite{achlioptas2020referit_3d}. We report accuracy at an IoU threshold of 0.1 (\%) for different query characteristics.
    \end{flushleft}}
    \label{tab:obj_ret_com}
\end{table*}

\subsection{Functional Elements 3D Segmentation}
\label{subsec:fun_sem_Seg}
To evaluate the segmentation performance of functional elements in \ours, we use the FunGraph3D dataset~\cite{zhang2025open}, a collection of annotated functional interactive elements within 3D scenes.
We compare \ours against FunGraph~\cite{rotondi2025fungraph} and OpenFunGraph~\cite{zhang2025open}, the only two 3DSG methods specifically designed for this task.
Our primary evaluation metric is Recall@K (R@K) for K=$\{3,5,10\}$ across all scenes in the dataset. A prediction is considered a true positive if the open-vocabulary class assigned to a segmented functional element has its embedding ranked within the top-k closest to the ground-truth class among the available labels, and if the IoU with the ground-truth segment exceeds a specified threshold.
We report results under different IoU thresholds $(0.0, 0.10, 0.25)$ in Tab.~\ref{tab:recall_results}.
\ours outperforms both FunGraph and OpenFunGraph across all metrics except Recall@10 with IoU$_{\ge0.0}$, highlighting that while OpenFunGraph tends to detect more functional elements, it is far less accurate in segmenting their geometry.

\subsection{Object Retrieval From Large-Scale Environment}
\label{subsec:obj_ret_hier}

To evaluate \ours's capabilities for retrieving objects in large-scale multi-floor environments, we utilized the Habitat Matterport 3D Semantic Dataset~\cite{habitat23semantics} , which features scenes with multiple floors and rooms. We selected RGB-D sequences from scenes \texttt{00824, 00829, 00843, 00861, 00862, 00873, 00877, 00890}, following the protocol in~\cite{Werby-RSS-24}. Following with~\cite{Werby-RSS-24}, we assessed performance on two query types: floor-room-object (e.g., ``the toilet in the bathroom on the ground floor'') and room-object (e.g., ``oven in the kitchen''). The evaluation spans 345 object categories with 2,809 queries per type.
As shown in Tab.~\ref{tab:obj_ret_hier}, we compared two variants of \ours against HOV-SG~\cite{Werby-RSS-24}. We report Recall@K (R@K) for $K \in \{1, 5, 10\}$. A retrieval is considered successful if the target object appears within the top-$K$ results and its Intersection over Union (IoU) with the ground truth meets or exceeds a specified threshold $\tau \in \{0.0, 0.10, 0.50\}$.
In the first variant (\ours w/o RAG), we relied on an LLM to decompose the hierarchical query into \texttt{[<floor>, <room>, <object>]}, mirroring the baseline~\cite{Werby-RSS-24}. We computed CLIP text embeddings for each parsed concept and hierarchically compared their cosine similarities with the corresponding CLIP embedding at each level of the scene graph.
In the second variant (\ours w/ RAG), we retrieved the multi-modal context $C^*$ by computing the posterior probability $P(C|q)$ given the query, as detailed in Sec.~\ref{subsec:rag}, utilizing the augmented text and image data for each graph level. We then selected the top-$k$ objects from the context $C^*$ and computed the IoU.
\ours outperformed HOV-SG~\cite{Werby-RSS-24} in both variants, with the RAG-based approach achieving competitive results.

\subsection{Object Grounding from Language Queries on Nr3D}
\label{subsec:obj_ret_com}

Finally, to evaluate \ours on grounding objects from language queries in indoor cluttered environments, we used the Nr3D dataset~\cite{achlioptas2020referit_3d}.
It provides a diverse set of natural language queries categorized into classes based on how the target object is referenced. Following the evaluation protocol of~\cite{linok2025beyondbarequeries}, we utilized scenes \texttt{0011\_00}, \texttt{0030\_00}, \texttt{0046\_00}, \texttt{0086\_00}, \texttt{0222\_00}, \texttt{0378\_00}, \texttt{0389\_00}, and \texttt{0435\_00}.
In Tab.~\ref{tab:obj_ret_com}, we report grounding accuracy at an IoU threshold of $0.1$. The results are presented overall and categorized by query characteristics, including the presence of spatial, color, or shape language, as well as explicit target mentions.

\begin{figure*}[t]
    \centering
    \includegraphics[height=12cm, keepaspectratio]{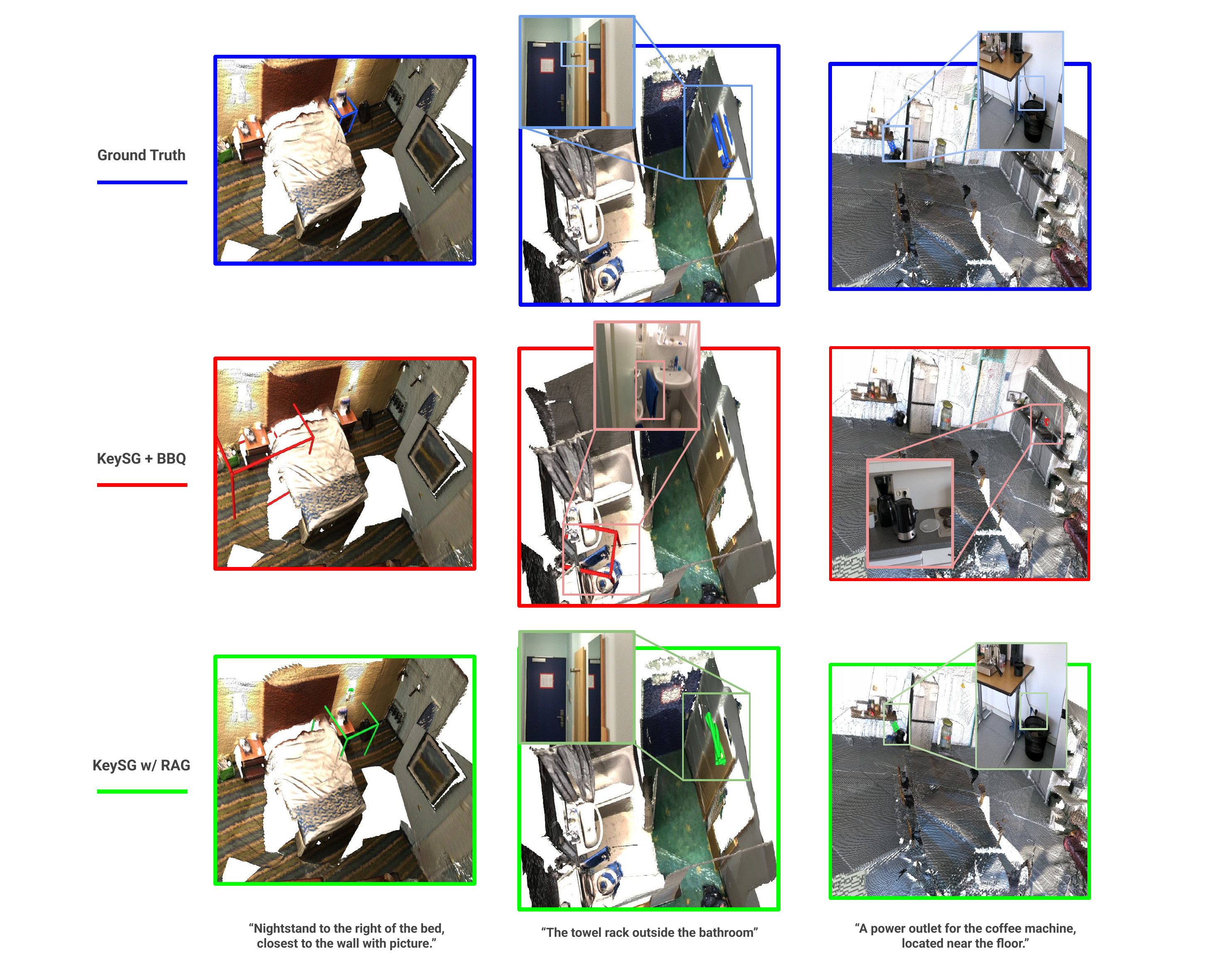}
    \vspace*{-.2cm}
    \caption{Qualitative illustration of \ours for grounding 3D objects from ambiguous queries. We compare our approach (\ours w/ RAG, bottom) against a baseline that relies on explicit, predefined spatial relations (KeySG w/ BBQ \cite{linok2025beyondbarequeries}, middle). The fixed-edge scene graph fails to find objects when queries contain multiple complex spatial cues or implicit boundaries. \ours w/ RAG successfully disambiguates objects as it leverages the implicit information stored in the text descriptions and raw images. These results demonstrate that relationships inferred on demand are superior in handling open-set, complex queries to edges that are rigidly predefined at construction time.}
    \label{fig:qualitative}
    \vspace*{-.2cm}
\end{figure*}

We compared two variants of \ours against recent scene graph methods~\cite{conceptgraphs, linok2025beyondbarequeries, yamazaki2024open}. The first variant, \ours w/ BBQ~\cite{linok2025beyondbarequeries}, constructs the \ours scene graph but replaces our RAG pipeline with a fixed set of semantic and metric relations between objects. The second variant, \ours w/ RAG, utilizes our full multi-modal RAG pipeline. As shown in Tab.~\ref{tab:obj_ret_com}, \ours w/ RAG outperforms all baselines relying on explicit spatial and semantic edges, including the \ours w/ BBQ variant, across all categories. This demonstrates the strict limitations of explicitly modeled relations, which often fail when queries contain complex spatial cues or implicit boundaries (see Fig.~\ref{fig:qualitative}). The performance gap is clearly visible in queries without spatial language (57.0\% vs. 43.0\%), showing that \ours w/ RAG effectively captures implicit contextual clues that fixed relational edges miss. Furthermore, \ours w/ RAG shows substantial gains in queries requiring fine-grained visual reasoning, such as shape (50.0\% vs. 28.9\%) and color (48.3\% vs. 46.5\%), highlighting the advantage of our multimodal RAG retrieving keyframes. Finally, while queries without an explicit target mention remain the most difficult category overall (25.4\%), \ours w/ RAG still achieves a relative improvement over the strongest baseline, showing robust deductive reasoning capabilities even when the semantic object tag is missing.

\begin{figure}[!t]
    \centering
    \includegraphics[width=\columnwidth]{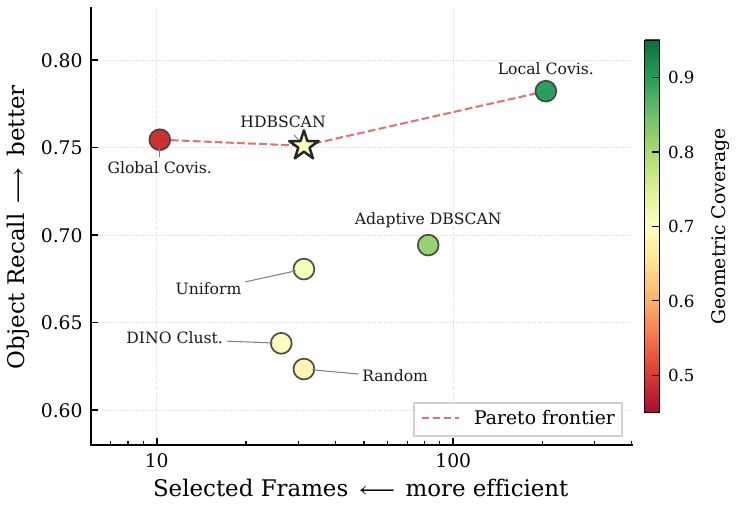}
    \caption{\textbf{Efficiency vs. Recall Frontier.} We compare keyframe-selection strategies based on three metrics defined in Sec.~\ref{subsec:ablation}. 
    \textbf{X-axis:} Selected frames (average frame count, log scale; left is more efficient). 
    \textbf{Y-axis:} Object Recall (visual proxy). 
    \textbf{Bubble Color:} Geometric Coverage (spatial proxy). 
    \textbf{HDBSCAN} achieves the optimal trade-off, attaining 96\% of the peak recall (found by Covisibility) while using $6.5\times$ fewer frames, avoiding the severe coverage drop observed in Covisibility Global.}
    \label{fig:ablation_frontier}
\end{figure}

\subsection{Ablation}
\label{subsec:ablation}
In order to evaluate the capability of different keyframe-selection strategies to preserve spatial and visual information while minimizing the number of keyframes, we design an ablation study comparing our proposed pose-based \textbf{HDBSCAN} \cite{hdbscan} clustering against six baseline strategies. To quantify the effectiveness of each method, we define three metrics: first, \textit{Object Recall}, defined as the percentage of unique ground-truth objects visible in the selected subset, serving as a proxy for preserving visual information; second, \textit{Geometric Coverage}, defined as the fraction of surface points in the full reconstruction within a distance threshold $\tau=0.05$m of the sparse keyframe-based point cloud, serving as a proxy for preserving spatial information; and finally, \textit{Sampling Ratio}, defined as the ratio of retained keyframes over the total number of available frames. We evaluate seven distinct sampling methods: naive \textit{Uniform} and \textit{Random} baselines; visual feature-based clustering leveraging \textit{DINOv2} embeddings \cite{oquab2023dinov2}; an \textit{Adaptive DBSCAN} approach that iteratively relaxes parameters to meet geometric coverage target; and two covisibility optimization strategies, a \textit{Local} sliding-window approach solving a set-cover problem, and a \textit{Global} greedy approach maximizing total object quality and redundancy.
In Fig \ref{fig:ablation_frontier}, we report the results for the ablation study. The results show that pose-based clustering with \textbf{HDBSCAN} \cite{hdbscan} is the sweet spot between \textit{Object Recall}, \textit{Geometric Coverage}, and \textit{Sampling Ratio}.

%% file: sections/5_conclusion.tex
\section{Limitations}

Our framework, while effective, has several limitations.
First, its reliance on computationally expensive large language and vision-language models makes graph construction an offline process that requires a pre-reconstructed scene. However, once the graph is built, it can be deployed on a robot as a persistent knowledge base that can be efficiently queried in real time thanks to the RAG pipeline.
Second, our method currently assumes a static environment and does not handle dynamic objects or changes in object states. These areas offer clear directions for future research.

\section{Conclusion}

In this work, we addressed a fundamental limitation of existing 3D Scene Graphs (3DSGs): their reliance on predefined relationship edges that restrict open-set semantic reasoning. We introduced \oursextended (\ours), the first framework to extend 3DSGs across multiple resolutions, ranging from full buildings to functional elements. By combining keyframe sampling, hierarchical scene summarization, and a retrieval-augmented generation mechanism, \ours efficiently grounds task-relevant context for large language models while overcoming traditional scalability limits. This enables the framework to handle a wide variety of open-ended tasks and queries. Finally, our experiments across open-vocabulary 3D segmentation, functional element extraction, and 3D object grounding benchmarks demonstrate that \ours outperforms prior approaches on most metrics.

\section{Acknowledgements}

\footnotesize{We would to like thank the anonymous reviewers for the constructive comments regarding the ablation study. This work has been supported by the German Federal Ministry of Research, Technology, and Space (BMFTR) under the Robotics Institute Germany (RIG).}
\vspace{-.1cm}